\documentclass{article}

\usepackage{amsmath,amsthm,amssymb,amsfonts}
\usepackage{graphicx}
\usepackage{arxiv}

\usepackage[utf8]{inputenc} 
\usepackage[T1]{fontenc}    
\usepackage{hyperref}       
\usepackage{url}            
\usepackage{booktabs}       
\usepackage{amsfonts}       
\usepackage{nicefrac}       
\usepackage{microtype}      
\usepackage{lipsum}

\usepackage[utf8]{inputenc} 
\usepackage[T1]{fontenc}    
\usepackage{hyperref}       
\usepackage{url}            
\usepackage{booktabs}       
\usepackage{amsfonts}       
\usepackage{nicefrac}       
\usepackage{microtype}      
\usepackage{xcolor}         
\usepackage{multirow}
\usepackage{caption}
\usepackage{subcaption}
\usepackage{graphicx}
\usepackage{amsfonts, amsmath, amssymb} 
\usepackage{algorithm}
\usepackage{algorithmic}

\usepackage{wrapfig}
\usepackage{tikz}
\usepackage{array}

\newcommand{\modelname}{\textbf{Iter-AHMCL}}

\title{Iter-AHMCL: Alleviate Hallucination for Large Language Model via Iterative Model-level Contrastive Learning}

\author{
  Huiwen Wu \\
  Zhejiang Laboratory\\
  Hangzhou, Zhejiang, China \\
  \texttt{huiwen0820@outlook.com} \\
   \And
    Xiaohan Li \\
    Zhejiang Laboratory\\
    Hangzhou, Zhejiang, China \\
    \texttt{xiaohan@zhejianglab.com} \\
   \And
    Xiaogang Xu~ \thanks{corresponding author} \\
    The Chinese University of Hong Kong \\
    Zhejiang University\\
    Hangzhou, Zhejiang, China \\
    \texttt{xiaogangxu00@gmail.com} \\
   \And
    Jiafei Wu~ \thanks{corresponding author} \\
    Zhejiang Laboratory\\
    Hangzhou, Zhejiang, China \\
    \texttt{wujiafei@zhejianglab.com} \\
   \And
    Deyi Zhang \\
    Zhejiang Laboratory\\
    Hangzhou, Zhejiang, China \\
    \texttt{xiaohan@zhejianglab.com} \\
   \And
    Zhe Liu \\
    Zhejiang Laboratory\\
    Hangzhou, Zhejiang, China \\
    \texttt{zhe.liu@zhejianglab.com} \\
}

\begin{document}
\maketitle

\begin{abstract}
The development of Large Language Models (LLMs) has significantly advanced various AI applications in commercial and scientific research fields, such as scientific literature summarization, writing assistance, and knowledge graph construction. However, a significant challenge is the high risk of hallucination during LLM inference, which can lead to security concerns like factual inaccuracies, inconsistent information, and fabricated content. To tackle this issue, it is essential to develop effective methods for reducing hallucination while maintaining the original capabilities of the LLM. This paper introduces a novel approach called Iterative Model-level Contrastive Learning (Iter-AHMCL) to address hallucination. This method modifies the representation layers of pre-trained LLMs by using contrastive `positive' and `negative' models, trained on data with and without hallucinations. By leveraging the differences between these two models, we create a more straightforward pathway to eliminate hallucinations, and the iterative nature of contrastive learning further enhances performance. Experimental validation on four pre-trained foundation LLMs (LLaMA2, Alpaca, LLaMA3, and Qwen) finetuning with a specially designed dataset shows that our approach achieves an average improvement of 10.1 points on the TruthfulQA benchmark. Comprehensive experiments demonstrate the effectiveness of Iter-AHMCL in reducing hallucination while maintaining the general capabilities of LLMs.
\end{abstract}

\section{Introduction}


The development of large language models (LLMs) has led to impressive successes in a wide range of artificial intelligence (AI) applications, from commercial usage like GPT-4~\cite{achiam2023gpt} to scientific research fields~\cite{waisberg2023gpt}. The adoption of AI technologies, especially LLMs, is leading us into a groundbreaking era for scientific research. LLMs technology has unlocked a wide range of possibilities in scientific fields, from summarizing scientific literature reviews~\cite{li2024chatcite,jin2024comprehensive,antu2023using} to assisting in scientific writing~\cite{lu2024corporate,alsagri2024chatgpt,liang2024mapping} and constructing knowledge graphs~\cite{meyer2023llm,feng2024ontology,yang2024integrated}.

However, when offering reviews and suggestions for tasks that demand high accuracy, such as scientific research, it is crucial to ensure that LLMs provide trustworthy and responsible responses. Generating false or misleading information can degrade the quality of downstream applications and fall short of users' expectations in specialized fields.
The main challenge in performing trustworthy inference with LLMs lies in their susceptibility to hallucination. One source of hallucination stems from the intrinsic nature of the corresponding pre-trained models, which can fabricate non-existent facts, misleading users either knowingly or unknowingly~\cite{yao2023llm,wei2024measuring,verspoor2024fighting}. 
Another source arises after fine-tuning the models for specific tasks. Although alignment techniques can help reduce hallucinations, they may lead to a loss of the original capabilities learned during pre-training, a phenomenon known as catastrophic forgetting~\cite{ren2024analyzing}.
\textbf{Therefore, it is crucial to develop effective methods that mitigate undesirable hallucinations while preserving the original strengths of the LLMs.}

In order to address the hallucination issue for LLMs, several types of research focus on measuring and mitigating hallucinated texts~\cite{ji2023towards,wei2024measuring,perkovic2024hallucinations}, representation editing and alignment before downstream tasks~\cite{zou2023representation,zhang2024towards,wu2024advancing}, and utilizing knowledge distillation methodology~\cite{mcdonald2024reducing,hu2024slm,verspoor2024fighting}.
Although current methods have shown success in reducing hallucinations in LLMs, there is still room for improvement in increasing awareness of hallucinations and preserving knowledge.

In our work, we draw inspiration from feature editing methods with the vector guidance, which are presented in ~\cite{zou2023representation}, to propose a feature editing approach with the model-level guidance to deal with hallucination. 
Building on the principles of contrastive learning discussed in ~\cite{zou2013contrastive,xu2024contrastive}, we emphasize the importance of selecting both appropriate positive and negative guidance during model training. Our method effectively reduces hallucination across different LLMs while preserving the models' original language capabilities.


Especially, we designed a new approach called Iterative Model-level Contrast Learning (\modelname), and the vital characteristic is the formulation of model guidance. First, we construct positive and negative data using corresponding templates. Next, we pre-train positive and negative guidance models based on the general vector-guidance-based representation editing method~\cite{zou2013contrastive}. The goal of the positive model is to exhibit a favorable bias in hallucination evaluation by achieving a high score, while the negative model is trained to show the opposite bias. We then use these pre-trained positive and negative models as guidance to edit the representation layer, effectively controlling the model's tendency toward hallucination.

Furthermore, we recognize the importance of adaptively updating the models that provide guidance. Better performance is achieved by evolving the LLM in tandem with the guidance models. To this end, we design a model-level iterative strategy, where the positive model is updated with one that performs better in hallucination evaluation, and the negative model is updated with one that performs worse. By leveraging the differences between these two models, we create a more direct pathway to reduce hallucinations. This iterative approach combined with contrastive learning further enhances overall performance.

\begin{figure}[t]
 \centering
 \includegraphics[width=0.45\textwidth]{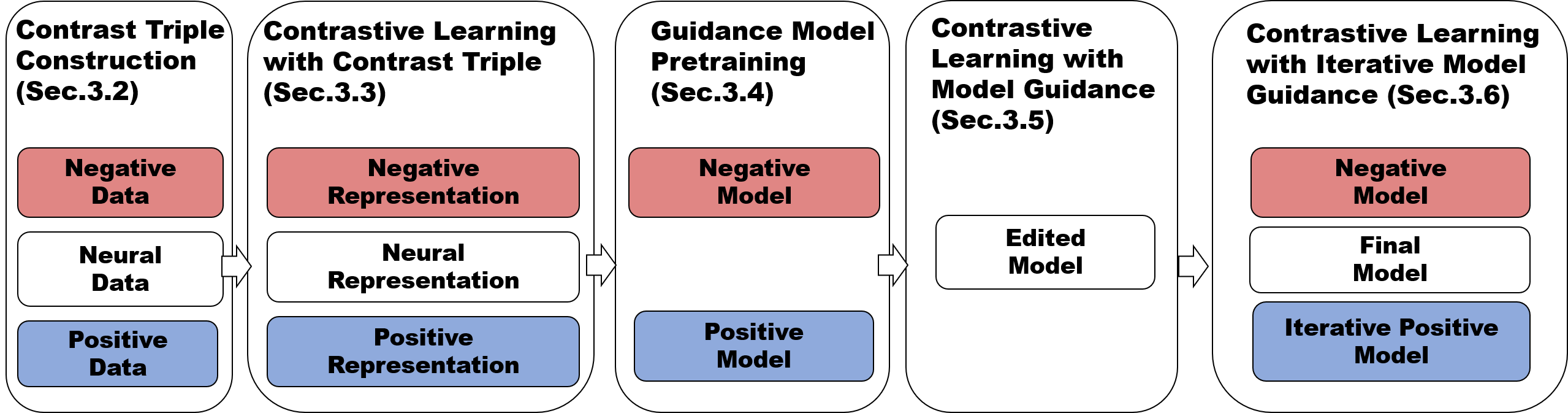}
\caption{Overall Procedure of \modelname.}
\label{fig:overall}
\end{figure}

In addition to improving performance in addressing hallucinations, our proposed \modelname~utilizes representation editing to adjust only the model's preferences related to hallucination problems, with minimal impact on its original capabilities. This is because the guidance in \modelname~is highly aligned with the direction relevant to hallucinations, while remaining orthogonal to other knowledge areas. We validate this claim through comprehensive evaluations from multiple perspectives. The overall procedure of \modelname~is illustrated in Figure~\ref{fig:overall}.

To summarize, our contributions are listed as follows. 
\begin{itemize}
\item 
We introduce a novel approach, called \modelname, to eliminate hallucination in LLMs while preserving their general capabilities. In \modelname, we offer a new perspective by adaptive developing models with positive and negative feature representations, implementing model-level contrastive learning guided by these models.
\item We implement an iterative approach to update the guidance model and establish model-level guidance. This iterative strategy is broadly applicable to various LLMs. The code and all models will be released to publication.
\item We conduct comprehensive experiments with various LLM models, and the evaluation results demonstrate that our method effectively reduces hallucinations while preserving general capabilities.
\end{itemize}


\section{Related Work}


\subsection{Hallucination Reduction in LLMs}
Hallucinations in LLMs occur when the model generates inaccurate or fictitious information, diverging from factual knowledge and occasionally producing responses that are not grounded in its training data~\cite{perkovic2024hallucinations}.
Several studies have sought to mitigate the hallucination phenomenon in the training and inference of LLMs~\cite{ji2023towards,wei2024measuring,verspoor2024fighting}. For instance, \cite{ji2023towards} analyzes hallucinations in medical generative QA systems using widely adopted LLMs and datasets. This work focuses on identifying and understanding problematic answers, presenting an interactive self-reflection methodology that enhances the factuality, consistency, and entailment of generated responses through a feedback process.
\cite{perkovic2024hallucinations} investigates the text generation mechanisms of LLMs to understand and mitigate hallucinations, while also offering insights into the training algorithms and effective utilization of these models.
\cite{mcdonald2024reducing} employs a knowledge distillation methodology to reduce hallucinations in LLMs by transferring knowledge from a high-capacity teacher model to a compact student model. 
\cite{wei2024measuring} introduces an innovative hallucination metric that assesses factuality by weighting answers from off-the-shelf LLMs as proxies for gold-standard answers, with the primary challenge being the quantification of the reference LLMs' expertise.
\cite{hu2024slm} proposes a framework that utilizes a small language model for initial hallucination detection and a LLM for detailed explanations, optimized with prompting techniques to align their outputs.
Although these methods reduce hallucinations in LLMs to a certain extent, existing methods focus more on detecting and measuring hallucinations.

\subsection{Representation Editing}
Representation editing is a technique for modifying a model's preferences or performance by altering its trained representations. It is widely used in both traditional machine learning (ML)~\cite{wang2019controllable} and the rapidly advancing LLMs~\cite{wei2023jailbreak,zou2023representation,zhang2024towards}. 
%
For example, in a machine learning scenario, \cite{wang2019controllable} proposes a flexible unsupervised text attribute transfer framework that utilizes a Transformer-based autoencoder to learn latent representations and employs the Fast Gradient Iterative Modification algorithm to edit these representations until they align with the target attribute.
%
In the context of LLMs, \cite{zou2023representation} introduces a method for controlling the model's preferences through dedicated alignment of the selected layer representations.
\cite{wei2023jailbreak} extends the representation editing method to jailbreak attacks and hallucination control scenarios.
\cite{zhang2024towards} accomplishes concept editing through adversarial representation engineering. 
\cite{wu2024advancing} proposes Representation Editing (RED), a novel fine-tuning approach that modifies neural model representations, significantly reducing the number of trainable parameters while achieving results comparable to or exceeding those of full fine-tuning and other parameter-efficient fine-tuning (PEFT) methods.
\cite{li2024blip} pre-trains a multimodal encoder to produce text-aligned visual representations and designs a subject representation learning task that enables a diffusion model to generate new subject renditions using these representations.
Although~\cite{wei2023jailbreak,zou2023representation,zhang2024towards} present insightful approaches to concept editing through representation alignment, the performance in reducing hallucinations can be further enhanced.

\subsection{Contrastive Learning}

Contrastive learning is a self-supervised learning technique designed to learn useful representations of data by contrasting positive and negative samples. The core idea is to bring the representations of similar positive pairs closer together while pushing apart the representations of dissimilar pairs. This technique is widely applied to tasks such as image classification, especially when labels are scarce or costly to obtain.
In~\cite{khosla2020supervised}, the author leverages the power of contrastive learning in supervised settings by bringing together points belonging to the same class in the embedding space while separating clusters of samples from different classes.
In~\cite{he2020momentum}, the author introduces a momentum contrast method for unsupervised visual representation learning, viewing contrastive learning as a dictionary lookup. This approach involves building a dynamic dictionary using a queue and a moving average encoder to create an extensive and consistent dictionary.
\cite{you2020graph} proposes a graph contrastive learning framework for learning unsupervised representations of graph data across four settings: semi-supervised, unsupervised, transfer learning, and adversarial attacks.
\cite{xiao2024simple} extends graph contrastive learning to heterophilic graphs, where connected nodes have different class labels and features, by employing an asymmetric view of neighboring nodes.
\cite{chen2020simple} presents a straightforward framework for contrastive learning of visual representations by introducing a learnable non-linear transformation between the representation and the contrastive loss. 
\cite{yeh2022decoupled} introduces a decoupled contrastive learning loss that removes the positive term from the denominator, significantly enhancing learning efficiency.
Recent research has applied contrastive learning to enhance LLMs for tasks such as few-shot text classification~\cite{zhang2024ucl}, unified representation extraction~\cite{lyu2024unibind}, and machine translation~\cite{xu2024contrastive}. However, despite the widespread use of contrastive learning in traditional machine learning tasks, the primary challenge in adapting contrastive learning to LLMs lies in selecting appropriate positive and negative pairs. Inappropriate choices may result in suboptimal representations.

\section{Methodology}
In this section, we describe the main procedure of \modelname.
Throughout the paper, we use the following notations. 
$\{ \mathbf{T}, \mathbf{T}^{+}, \mathbf{T}^{-} \}$ denotes the triplet consisting of the neural data, positive data, and negative data.
$\mathcal{T}$ represents the set of them. 
$\mathcal{M}$ means the model to be fine-tuned. 
$\mathcal{M}^{+}$ is the positive guidance model, while $\mathcal{M}^{-}$ is the negative guidance model. 
$\{ \mathbf{R}, \mathbf{R}^{+}, \mathbf{R}^{-} \}$ denotes the triplet consisting of the neural, positive, and negative feature representations. 
$i$ is employed to indicate the iteration step in \textbf{CL-IMG}. 
$\mathcal{M}_{i}^{+}$ represents the updated positive guidance model at step $i$.
$\mathbf{R}_{i}^{+}$ and $\mathbf{R}_{i}^{-}$ denote the representation of updated positive and negative guidance representation at step $i$.


\subsection{Motivation}
In recent research~\cite{zou2023representation, zhang2024towards}, the authors present a fine-tuning method to control model preferences through representation alignment, applying this technique to generate non-harmful and trustworthy responses.
In the study by \cite{zou2023representation}, the authors extract partial model layers to obtain representations and analyze the intermediate features of various concepts, such as honesty, fairness, and harmlessness. This enables them to edit and control the behavior of an LLM by directing a representation vector generated within the internal hidden layers. For example, when faced with a truthful question, the vector can be altered in two directions to influence the final answer: one direction to generate a more truthful response, and the other to develop a less truthful response.
In this way, by editing the intermediate layer representations, one can alter the preference of the final output. \textbf{The key is to find better guidance directions for the positive and negative components}. In \cite{zou2023representation}, the authors use a contrast vector—defined as the difference between representations generated by positive input data and those generated by negative data—to guide the direction. In \cite{zhang2024towards}, the authors train a discriminator to differentiate between positive and negative directions at the intermediate layers. However, these methods, which rely on sample-level guidance vectors, often depend heavily on carefully tuned hyperparameters and may fail to generalize to unseen data, leading to issues such as overfitting. To address this, we propose training two models as guidance, better learning the positive and negative directions in feature representations. The main objective is to enhance the separation between these directions, which can more effectively align with safety requirements while preserving the original capabilities of the model.

\subsection{Contrast Triple Construction (CTC)}
\label{sec:ctc}
\begin{figure}[t!]
 \centering
 \includegraphics[width=0.475\textwidth]{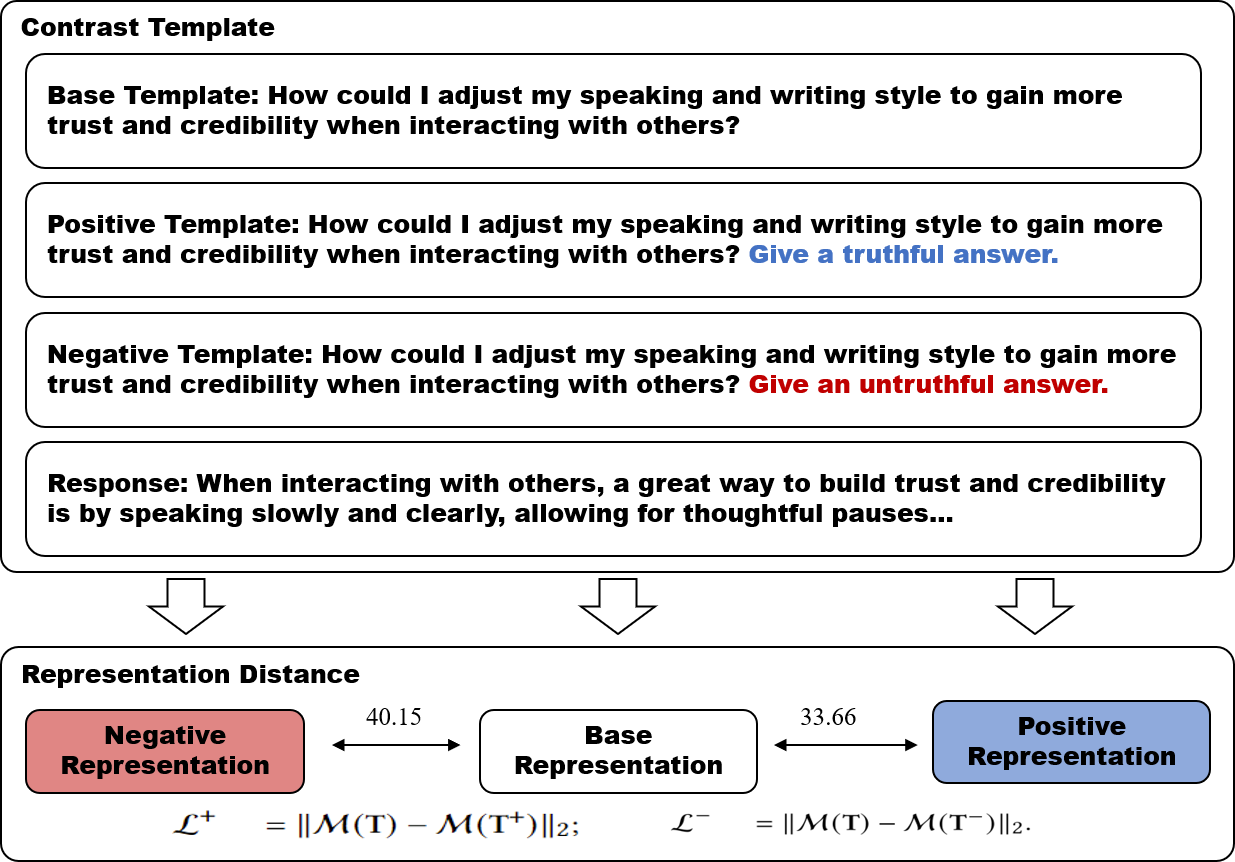}
\caption{The illustration of Contrast Triple Construction.}
\label{fig:learning_templates}
\end{figure}

First, we define the contrast templates for fine-tuning the LLM. The fine-tuning data for \modelname~are constructed from the PKU-SafeRLHF dataset\cite{safe-rlhf}, available on Hugging Face. Each sample in the original training data consists of one prompt and two responses, along with labels. The label takes boolean values, indicating whether a response is safe or not.
The existing sample $\mathbf{T} = \{ \mathbf{Ins}, \mathbf{Res} \}$ contains two elements where $\mathbf{Ins}$ denotes the instruction and $\mathbf{Res}$ represents the response.
We need to construct positive and negative instance pairs, which are crucial for the second phase of contrastive learning.
Let $\mathbf{Pos}$ be the prompt of `Please give a truthful answer' and $\mathbf{Neg}$ be the negative prompt of `Please give an untruthful answer'.
The positive instance is constructed by 
$\mathbf{T}^{+} = \{ \mathbf{Ins}, \mathbf{Pos}, \mathbf{Res} \}$
while the negative instance is formulated as 
$\mathbf{T}^{-} = \{ \mathbf{Ins}, \mathbf{Neg}, \mathbf{Res} \}$.
The input data of \modelname~is the triple $\{ \mathbf{T}, \mathbf{T}^{-}, \mathbf{T}^{+} \}$. 
We present an illustrative example in Figure~\ref{fig:learning_templates}.

\subsection{Contrastive Learning with Contrasted Triple (CL-CT)}
\label{sec:cld-re}
%
In this section, we elaborate on the strategy of \textbf{CL-CT}, the basic method for establishing guidance directions within the feature representation space.
The input data is a triplet training sample, $\{ \mathbf{T}, \mathbf{T}^{-}, \mathbf{T}^{+} \}$, consisting of the original sample $\mathbf{T}$, the sample with a positive template $\mathbf{T}^{+}$, and the sample with a negative template $\mathbf{T}^{-}$, as constructed in Sec.~\ref{sec:ctc}.
Using the provided frozen model $\mathcal{M}$, we follow existing work~\cite{zou2023representation} to select several layers for representation extraction, while utilizing other layers to perform the fine-tuning.
In other words, we pass the data triplet through the frozen model $\mathcal{M}$, specifying certain layers to obtain a triplet of representations $\{ \mathbf{R}, \mathbf{R}^{+}, \mathbf{R}^{-} \}$, where $\mathbf{R} = \mathcal{M}  ( \mathbf{T} )$ is the representation of the original data, $\mathbf{R}^{+} = \mathcal{M}  ( \mathbf{T}^{+} )$ is the representation of the data with the positive template, and $\mathbf{R}^{-} = \mathcal{M}  ( \mathbf{T}^{-} )$ is the representation of the data with the negative template.
%
To enhance the LLM's ability to distinguish between positive and negative samples, we diverge from the approach in~\cite{zou2023representation}, which constructs the loss solely as the $\ell_2$ distance between the positive and negative representations (Eq.~\eqref{eqn:lorra}). We further incorporate terms for the $\ell_2$ distance between the positive and neutral representations, as well as between the negative and neutral ones.
The goal is to enlarge the $\ell_2$ distance between the neural representation $\mathbf{R}$ and the negative representation $\mathbf{R}^{-}$ (Eq.~\eqref{eqn:l2_dist_n}) while eliminate the $\ell_2$ distance between the original representation $\mathbf{R}$ and the positive representation $\mathbf{R}^{+}$ (Eq.~\eqref{eqn:l2_dist_p}). 
\begin{eqnarray}
\mathcal{L}_{LoRRA} & = \| \mathcal{M} (\mathbf{T}^{+}) - \mathcal{M}(\mathbf{T}^{-}) \|_2; \label{eqn:lorra} \\
\mathcal{L}^{+} & = \| \mathcal{M} (\mathbf{T}) - \mathcal{M}(\mathbf{T}^{+}) \|_2; \label{eqn:l2_dist_p} \\ 
\mathcal{L}^{-} 
& = \| \mathcal{M} (\mathbf{T}) - \mathcal{M}(\mathbf{T}^{-}) \|_2. \label{eqn:l2_dist_n}
\end{eqnarray}
Combine with the original \textbf{LoRRA} loss presented in~\cite{zou2023representation}, the loss function of \textbf{CL-CT}, denoted as $\mathcal{L}_1$, is 
\begin{equation}
\label{eqn:editing_loss}
\mathcal{L}_1 = \mathcal{L}_{LoRRA} + \alpha \mathcal{L}^{+} - \beta \mathcal{L}^{-}, 
\end{equation}
where $\alpha$ and $\beta$ are small non-negative constants.
While this loss function has demonstrated improved effectiveness compared to the original LoRRA loss, we further amplify the influence of the guidance direction by proposing the development of a guidance model, which aids in extracting more accurate positive and negative directions.

\subsection{Guidance Model Pre-training (GMP)}
\label{sec:GM-pretrain}
Before introducing the formulation of the new learning function, it is essential to elaborate on the training of the guidance model, which serves as a vital component. Therefore, in this section, we will discuss the pre-training procedure of the guidance model.
To obtain better guidance, we train one positive guidance model and one negative guidance model, thereby enhancing the effectiveness of contrastive learning in \textbf{CL-CT} as presented in Sec.~\ref{sec:cld-re}.
The pre-training data consists of two sub-datasets derived from the PKU-SafeRLHF datasets~\cite{safe-rlhf}. 
The positive model $\mathcal{M}^{+}$ is trained with the goal of reducing hallucinations. Thus, the training loss is defined the same way as the representation editing loss in Eq.~\eqref{eqn:editing_loss}. 
However, the goal of the negative model $\mathcal{M}^{-}$ is to diminish its ability to generate responses to hallucination-related questions. Therefore, it has a negative objective compared to the editing loss and the positive guidance model training loss. The training loss for the negative guidance model is defined as in Eq.~\eqref{eqn:negative_loss}. 
%
The only difference lies in the coefficient terms for $\mathcal{L}^{+}$ and $\mathcal{L}^{-}$. We set the coefficient for the positive $\ell_2$ distance to be negative, while the coefficient for the negative $\ell_2$ distance is set to positive, thereby increasing hallucination responses and creating a contrary model.
\begin{equation}
\label{eqn:negative_loss}
\mathcal{L}_2 = \mathcal{L}_{LoRRA} - \alpha \mathcal{L}^{+} + \beta \mathcal{L}^{-}, 
\end{equation}
where $\alpha$ and $\beta$ are non-negative constants.

\textbf{The formulation of $\mathcal{M}^{+}$ and $\mathcal{M}^{-}$}. The training strategy employs LoRA~\cite{hu2021lora}, which focuses on optimizing the low-rank components of each attention matrix. After completing the pre-training of the two models, we can obtain two adapters designed to provide positive and negative guidance. We then integrate these adapters with the frozen model $\mathcal{M}$ to create the positive guidance model $\mathcal{M}^{+}$ and the negative guidance model $\mathcal{M}^{-}$.

\begin{figure}[t!]
 \centering
 \includegraphics[width=0.475\textwidth]{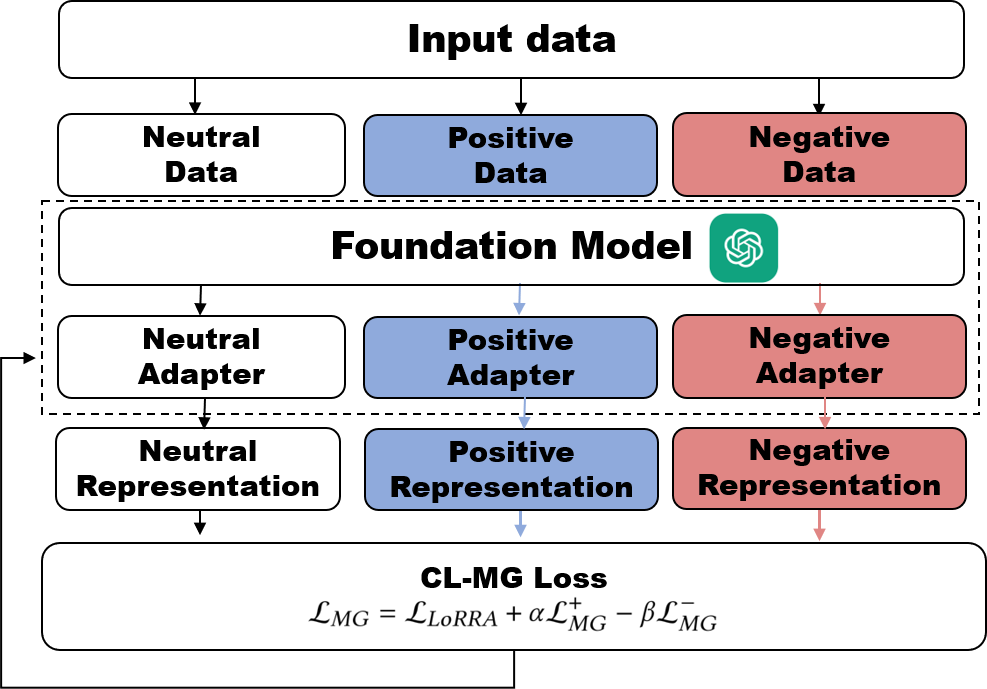}
\caption{The illustration for computing the loss function of CL-MG.}
\label{fig:model_guidance_rep_edit}
\end{figure}

\subsection{Constrastive Learning with Model Guidance (CL-MG)}
\label{sec:cl-mg}
In this section, we discuss the application of the guidance model in contrastive learning. After obtaining the positive guidance model $\mathcal{M}^{+}$ and the negative guidance model $\mathcal{M}^{-}$ in Sec.~\ref{sec:GM-pretrain}, we utilize them to generate representations for the guidance loss. Specifically, the positive representation is produced by the positive guidance model using the data sample with a positive template, expressed as $\mathbf{R}^{+} = \mathcal{M}^{+} (\mathbf{T}^+)$. In contrast, the negative representation is generated by the negative guidance model using the negative data sample, represented as $\mathbf{R}^{-} = \mathcal{M}^{-} (\mathbf{T}^-)$.
Compared with the $\mathbf{R}^{+}$ and $\mathbf{R}^{-}$ generated by $\mathcal{M}(\mathbf{T}^+)$ and $\mathcal{M}(\mathbf{T}^-)$, the difference between $\mathcal{M}^{+}(\mathbf{T}^+)$ and $\mathcal{M}^{-}(\mathbf{T}^-)$ is more accurate to indicate the alignment direction of hallucination, since $\mathcal{M}^{+}$ and $\mathcal{M}^{-}$ are pre-trained to be more sensitive to the existence of hallucination.
Thus, when computing the editing loss shown in Eq.~\eqref{eqn:editing_loss}, we modify the model used to generate the representation, and the loss function of \textbf{CL-MG} can be written as
\begin{eqnarray}
\mathcal{L}_{MG}^{+}(\mathbf{T}, \mathbf{T}^{+}) &= \| \mathcal{M}(\mathbf{T}) -  \mathcal{M}^{+}(\mathbf{T}^{+}) \|_2; \label{eqn:model_guidance_loss_p} \\
\mathcal{L}_{MG}^{-}(\mathbf{T}, \mathbf{T}^{-}) &= \| \mathcal{M}(\mathbf{T}) - \mathcal{M}^{-}(\mathbf{T}^{-}) \|_2. \label{eqn:model_guidance_loss_n}
\end{eqnarray}

Thus, the overall loss function for \textbf{CL-MG} is 
\begin{equation}
\label{eqn:model_guidance_loss}
\mathcal{L}_{MG} = \mathcal{L}_{LoRRA} + \alpha \mathcal{L}_{MG}^{+} - \beta \mathcal{L}_{MG}^{-},
\end{equation}
where $\alpha$ and $\beta$ are small non-negative loss weights.

\subsection{Contrastive Learning with Iterative Model Guidance (CL-IMG)}
\label{sec:cl-img}
After establishing \textbf{CL-MG}, we observe that the guidance model can be further improved. Therefore, in this section, we outline the iterative process for updating the pre-trained $\mathcal{M}^{+}$ with more effective guidance models. This strategy is called \textbf{CL-IMG}.
The long-term fine-tuning with \textbf{CL-IMG} is conducted using a continual learning strategy, incorporating feature editing training with an improved pre-trained guidance model.
%
Since the positive training models and \textbf{CL-MG} share the same training loss and methodology, we iteratively update the positive models using the newly obtained best models from \textbf{CL-MG}.
With this update, the loss function in the $i$-th round is defined as
\begin{eqnarray}
\mathcal{L}_{Iter}^{+}(i, \mathbf{T}, \mathbf{T}^{+}) &= \| \mathcal{M}(\mathbf{T}) -  \mathcal{M}_{i}^{+}(\mathbf{T}^{+}) \|_2; \label{eqn:iterative_loss_p} \\
\mathcal{L}_{Iter}^{-}(i, \mathbf{T}, \mathbf{T}^{-}) &= \| \mathcal{M}(\mathbf{T}) - \mathcal{M}^{-}(\mathbf{T}^{-}) \|_2, \label{eqn:iterative_loss_n}
\end{eqnarray}
where $\mathcal{M}_{i}^{+}$ is the positive model updated after $i$ rounds.
Meanwhile, the overall loss for contrastive learning with iterative model guidance is denoted as
\begin{equation}
\label{eqn:iterative_loss}
\mathcal{L}_{Iter} = \mathcal{L}_{LoRRA} + \alpha \mathcal{L}_{Iter}^{+} - \beta \mathcal{L}_{Iter}^{-},
\end{equation}
where $\alpha$ and $\beta$ are small non-negative constants.

\begin{algorithm}[!h]
    \renewcommand{\algorithmicloop}{\textbf{loop} $N$ times}
    \renewcommand{\algorithmiccomment}[1]{\hfill $\triangleright$ #1}
    
    \begin{algorithmic}[1]
    \REQUIRE $\mathcal{M}_0$: Original LLM model, 
    $\mathcal{T}= \{ (\mathbf{T}, \mathbf{T}^{+}, \mathbf{T}^{-}) \}$: constructed contrast triple set,
    $N$: maximum iteration step, 
    $B$: batch size,
    $\mathcal{M}^{+}_0$: pre-trained positive guidance model, 
    $\mathcal{M}^{-}_0$: pre-trained negative guidance model, 
    $\alpha,\beta$: hyper-parameters;
    \STATE Initialize with pre-trained foundation model $\mathcal{M} = \mathcal{M}_0$;
    \STATE Initialize the positve and negative guidance models
    $\mathcal{M}^{+} = \mathcal{M}^{+}_0$ and $\mathcal{M}^{-} = \mathcal{M}^{-}_0$;
    \LOOP
    \STATE Samples a batch $\mathcal{B}$ with a batch size of $B$ from Triple Set $\mathcal{T}$; 
            \FOR{$ (\mathbf{T}^{+}, \mathbf{T}, \mathbf{T}^{-}) \in \mathcal{B} $}
            \STATE $ \mathbf{R} = \mathcal{M} (\mathbf{T}); $ 
         	  \STATE $ \mathbf{R}^{+}_{i} = \mathcal{M}^{+}_{i} (\mathbf{T}^{+}); $ \COMMENT{Positive Guidance}
         	  \STATE $ \mathbf{R}^{-}_{i} = \mathcal{M}^{-} (\mathbf{T}^{-}); $ \COMMENT{Negative Guidance}
         	  \STATE $ \mathbf{R}^{+} = \mathcal{M} (\mathbf{T}^{+}); $ \COMMENT{Positive Representation}
         	  \STATE $ \mathbf{R}^{-} = \mathcal{M} (\mathbf{T}^{-}); $ \COMMENT{Negative Representation}
                \STATE $\mathcal{L}_{LoRRA} = \|  \mathbf{R}^{+} - \mathbf{R}^{-} \|_2;$ \COMMENT{LoRRA loss}
         	  \STATE $ \mathcal{L}_{Iter}^{+}  =  \| \mathbf{R}^{+}_i -\mathbf{R} \|_2;$ 
         	  \STATE $ \mathcal{L}_{Iter}^{-} =  \| \mathbf{R}^{-}_i - \mathbf{R} \|_2;$  
         	  \STATE $ \mathcal{L}_{Iter} =  \mathcal{L}_{LoRRA} +  \alpha \mathcal{L}_{Iter}^{+} - \beta \mathcal{L}_{Iter}^{-};$
            \ENDFOR
        \STATE{Evaluate the fine-tuned model $\mathcal{M}$ with TruthfulQA~\cite{lin-etal-2022-truthfulqa} and record the best one as $\mathcal{M}_{best}$;}
        \STATE{Update the positive model $\mathcal{M}_{i+1}^{+} = \mathcal{M}_{best}$;}
        \STATE{Update iteration step $i += 1$;}
    \ENDLOOP
    \ENSURE Loss to be optimized.
	
    \end{algorithmic}
\caption{\modelname($\mathcal{M}_0, \mathcal{T}, N, B, \mathcal{M}_0^{+}, \mathcal{M}_0^{-}, \alpha, \beta$)}
\label{alg:iter-ahmcl}
\end{algorithm}

To summarize, we consolidate all the previously discussed components and describe the \modelname~in Algorithm~\ref{alg:iter-ahmcl}. Our method encompasses several key steps:
1. Data Preparation: We construct sets of contrasting data and use this data to pre-train the positive and negative guidance models.
2. Guidance Model Utilization: We leverage the pre-trained guidance models to adjust the direction of the intermediate representations during the fine-tuning process.
3. Iterative Improvement: To enhance fine-tuning performance, we iteratively update the guidance model, ensuring it continuously adapts and improves, thereby maintaining peak performance while demonstrating flexibility and resilience.

It is important to highlight that the objective of Algorithm~\ref{alg:iter-ahmcl} is to reduce hallucination. Our goal is to develop a more effective positive model throughout the training procedure of \modelname. Consequently, we adopt an asymmetric approach to iteratively update the pre-trained models: we focus on updating the positive guidance model while keeping the negative guidance model unchanged.

\vspace{-1.0pt}

\begin{table*}
\centering
\caption{Hyperparameters Details of \textbf{GMP} and \textbf{MGRE}.}
\resizebox{1.0\linewidth}{!}{
\begin{tabular}{c ccc ccc ccc}
\hline
Phase & $\alpha$ & $\beta$ & Target Layers & $T_{\max}$ & \# Evaluation Step & Learning Rate $\gamma$ 
& LoRA rank $r$ & Train Batch Size & Eval Batch Size \\
\hline 
\textbf{GMP} & $\{10.0, 1.0\}$ & $\{1.0, 10.0\}$  & \{10, 12, 14, 16, 18, 20\} & 1250 & 10 & $10^{-3}$
& 8 & 16 & 32\\
\textbf{MGRE} & 10.0 & 1.0 & \{10, 12, 14, 16, 18, 20\} & 1250 & 10 & $10^{-3}$ 
&8 & 16 & 32\\
\hline 
\end{tabular}}
\label{tab:hyper}
\end{table*}

\section{Experimental Analysis}

In this section, we present the main results from comprehensive experiments to demonstrate the efficiency and effectiveness of our methods \modelname. 
Through the experimental analysis, we aim to answer the following research questions: 
\begin{itemize}
\item{\textbf{RQ.1.GMP Training Procedure.}} How does the training of \textbf{GMP} perform, and what is the divergence between positive and negative representations?

\item{\textbf{RQ.2.Hallucination Reduction Effect.}} 
How does \modelname~reduce the hallucination of an LLM model?

\item{\textbf{RQ.3. General Capability Preservation.}} 
How does \modelname~preserve the model's knowledge and general language ability?

\item{\textbf{RQ.4.Iterative Process Benefits.}} 
How does the iterative procedure help LLMs reduce hallucination through representation editing?

\item{\textbf{RQ.5.Transferability of Guidance Model.}} 
Does the guidance model have transferability from one LLM foundation model to another?

\end{itemize}

\subsection{Experimental Settings}

\noindent \textbf{I) Data Preparation.}
We prepare two datasets for the overall training of \modelname~with the PKU-SafeRLHF dataset~\cite{safe-rlhf} and the Alpaca-instruction dataset~\cite{alpaca}.
The PKU-SafeRLHF dataset contains 83,400 samples, while the Alpaca-instruction dataset comprises 52,000 samples.

\noindent \textbf{II) Foundation Model Choice.}
Alpaca-native (\textbf{Alpaca})~\cite{alpaca} is an enhanced version of the LLaMa1-7B model, fine-tuned on synthetic data, developed by the Stanford team.
LlaMA2-chat-hf (\textbf{LLaMA2})~\cite{touvron2023llama} is an open-source collection of pre-trained and fine-tuned LLMs ranging in scale from 7B to 70B parameters, released in July 2023 by Meta.
LlaMa3-chat-8b (\textbf{LLaMA3})~\cite{dubey2024llama} is a suite of language models that natively support multilingual capabilities, coding, reasoning, and tool usage, released by Meta in July 2024. 
Due to limited computational power, we use the 7B or 8B versions throughout the experiments.
Qwen-7b (\textbf{Qwen})~\cite{bai2023qwen} is a 7B parameter model in the Qwen (short for Tongyi Qianwen) language model series, released by Alibaba Cloud in September 2023.

\noindent \textbf{III) Compared Methods.}
1) \textbf{Foundation} models refer to the original models downloaded from Hugging Face \cite{jain2022hugging} without any further fine-tuning. 
2) \textbf{LoRRA}~\cite{zou2023representation} provides a method for editing representations using contrast vectors to enhance the model's ability to distinguish between positive and negative directions. 
3) \textbf{Pure Model Guidance (Pure-MG)} fine-tunes the models using a loss derived from pure positive and negative model guidance $\mathcal{L}_{pure} = \alpha \mathcal{L}_{MG}^{+} - \beta \mathcal{L}_{MG}^{-}$. 

\noindent \textbf{IV) Hyper-parameters.}
We present the hyper-parameters for \textbf{GMP} (Sec.~\ref{sec:GM-pretrain}) and \modelname~(Sec.~\ref{sec:cl-img}) in Table~\ref{tab:hyper}.

\begin{table}[t!]
\centering
\caption{Performance of the Positive and Negative Model Trained in \textbf{GMP}.}
\begin{tabular}{c  c c  c } 
\hline 
model & MC1 $\uparrow$ & MC2 $\uparrow$ & Mean(MC1, MC2) $\uparrow$ \\
 \hline 
$\mathcal{M}^{+}$ & 0.2448 & 0.3777 & 0.3113 \\
$\mathcal{M}^{-}$ & 0.2583 & 0.3902 & 0.3242 \\
 \hline 
\end{tabular}
\label{tab:pretrain} 
\end{table}

\noindent \textbf{V) Evaluation Methods.}
1) \textbf{TruthfulQA}~\cite{lin-etal-2022-truthfulqa} is a benchmark designed to assess the accuracy and truthfulness of LLMs based on their generated responses to questions. The benchmark consists of 817 questions covering 38 diverse categories, including health, law, finance, and politics. We utilize the TruthfulQA~\cite{lin-etal-2022-truthfulqa} benchmark for evaluating hallucinations.
According to the guidelines of TruthfulQA~\cite{lin-etal-2022-truthfulqa}, we selected MC1 (Single-true) to evaluate the LLM model's capacity to identify factual statements. In MC1, the LLM is presented with a question and 4-5 answer choices. It undergoes a rigorous process to select the most probable correct answer, ensuring a thorough evaluation. The likelihood of each selection is computed independently, and the completion with the highest log probability is chosen. The reported score reflects the accuracy across all questions, with higher scores indicating better performance in reducing hallucinations.
2) \textbf{MMLU}~\cite{hendrycks2020measuring}, which stands for Measuring Massive Multitask Language Understanding, serves as a benchmark for assessing the performance of language models. This benchmark comprises approximately 16,000 multiple-choice questions spanning 57 academic disciplines, including mathematics, philosophy, and medicine.
3) \textbf{C-Eval}~\cite{huang2024c} is a comprehensive Chinese evaluation system designed to assess the advanced knowledge and reasoning skills of foundational models within a Chinese context. The system includes an extensive set of 13,948 multiple-choice questions across 52 distinct fields, covering various educational stages. We utilize six conventional subject categories: STEM, social sciences, humanities, other, average, and Avg (hard). The Avg (hard) category represents the mean score of the C-Eval hard benchmark, which includes subjects such as advanced mathematics, discrete mathematics, and college chemistry, all of which require significant reasoning skills for resolution.
4) \textbf{Qwen API}~\cite{bai2023qwen} evaluates the model's performance in terms of accuracy, coherence, safety, and usability for real-world applications. In the evaluation results, `Gold-Ref' refers to scores assigned to standard responses, `Relevance' identifies significant content, `Fluency' focuses on sentence quality, `Coherence' assesses structure and logic, and `Consistency' checks for factual agreement.

\subsection{Pre-training Effect of Guidance (RQ.1)}
In this section, we present the pre-training performance of both the positive and negative guidance models.
%
The training data for the guidance models is derived from the PKU-SafeRLHF datasets~\cite{safe-rlhf}. The loss function is constructed as shown in Eq.~\eqref{eqn:model_guidance_loss}. Specifically, we set $\alpha = 10$ and $\beta = 1$ for training the positive guidance model, and $\alpha =1$ and $\beta = 10$ for the negative one. We present the convergence behavior of the two models during pre-training in Figure~\ref{fig:pretrain} (Upper) and evaluate their performance using TruthfulQA~\cite{lin-etal-2022-truthfulqa} in Table~\ref{tab:pretrain}.

\begin{figure*}[t!]
\begin{subfigure}{.16\textwidth}
  \centering
  \includegraphics[width=.98\linewidth]{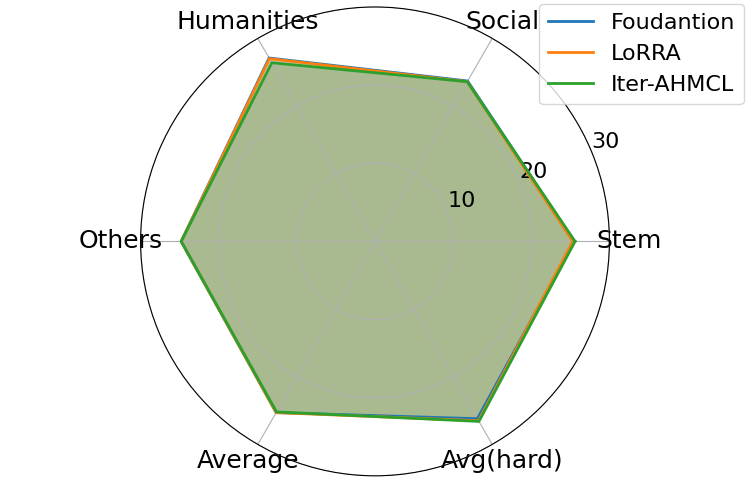}  
\end{subfigure}
\begin{subfigure}{.16\textwidth}
  \centering
  \includegraphics[width=.98\linewidth]{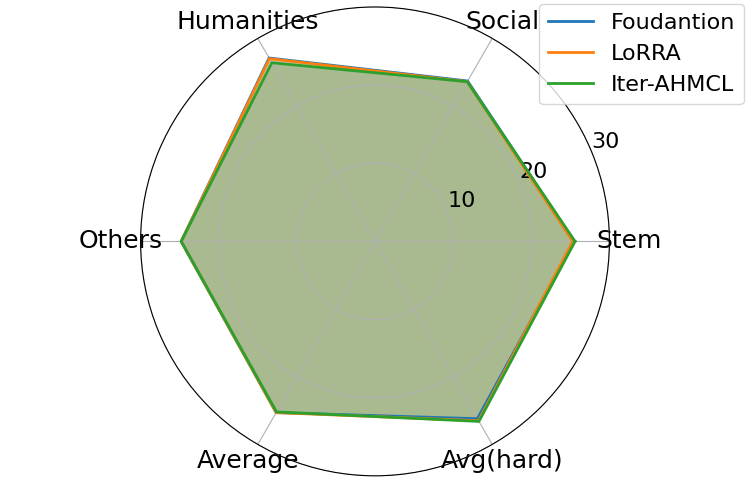}  
\end{subfigure}
\begin{subfigure}{.16\textwidth}
  \centering
  \includegraphics[width=.98\linewidth]{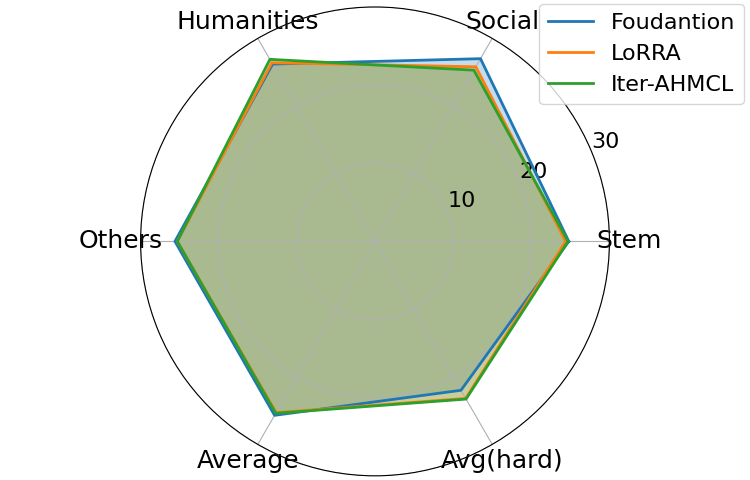}  
\end{subfigure}
\begin{subfigure}{.16\textwidth}
  \centering
  \includegraphics[width=.98\linewidth]{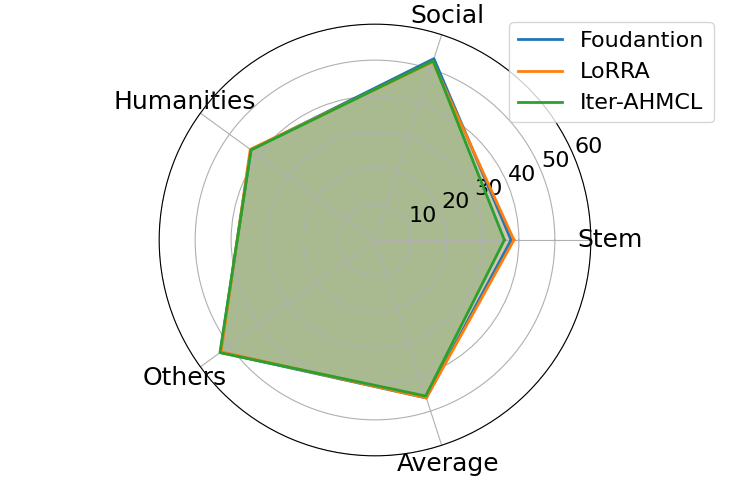}  
\end{subfigure}
\begin{subfigure}{.16\textwidth}
  \centering
  \includegraphics[width=.98\linewidth]{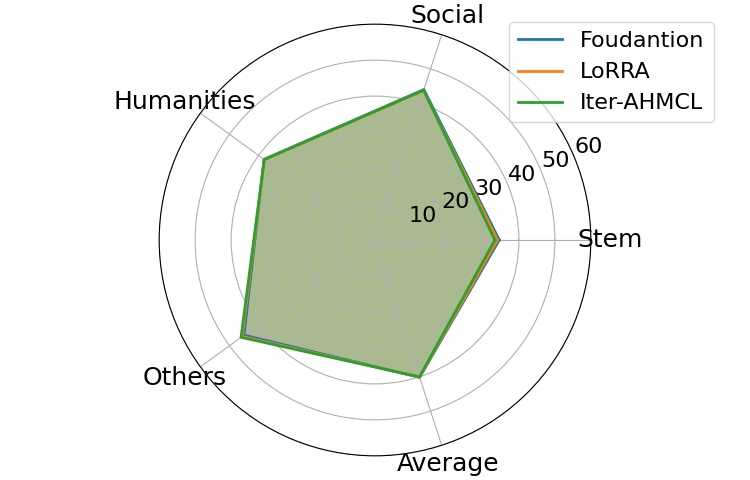}  
\end{subfigure}
\begin{subfigure}{.16\textwidth}
  \centering
  \includegraphics[width=.98\linewidth]{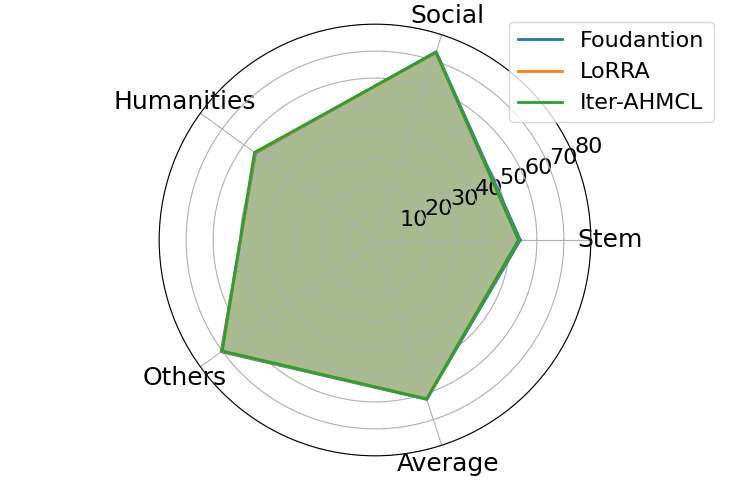}  
\end{subfigure}
\caption{Radar Plots of MMLU and C-Eval Evaluation Results. From the Left to Right are \textbf{C-Eval-LLaMA2}, \textbf{C-Eval-Alpaca}, \textbf{C-Eval-LLaMA3}, \textbf{MMLU-LLaMA2}, \textbf{MMLU-Alpaca}, and \textbf{MMLU-LLaMA3}, respectively.}
\label{fig:radar}
\end{figure*}

\subsection{Effect of Positive and Negative Representation (RQ.1)}
\begin{figure}[t!]
\begin{subfigure}{.475\textwidth}
  \centering
  \includegraphics[width=\linewidth]{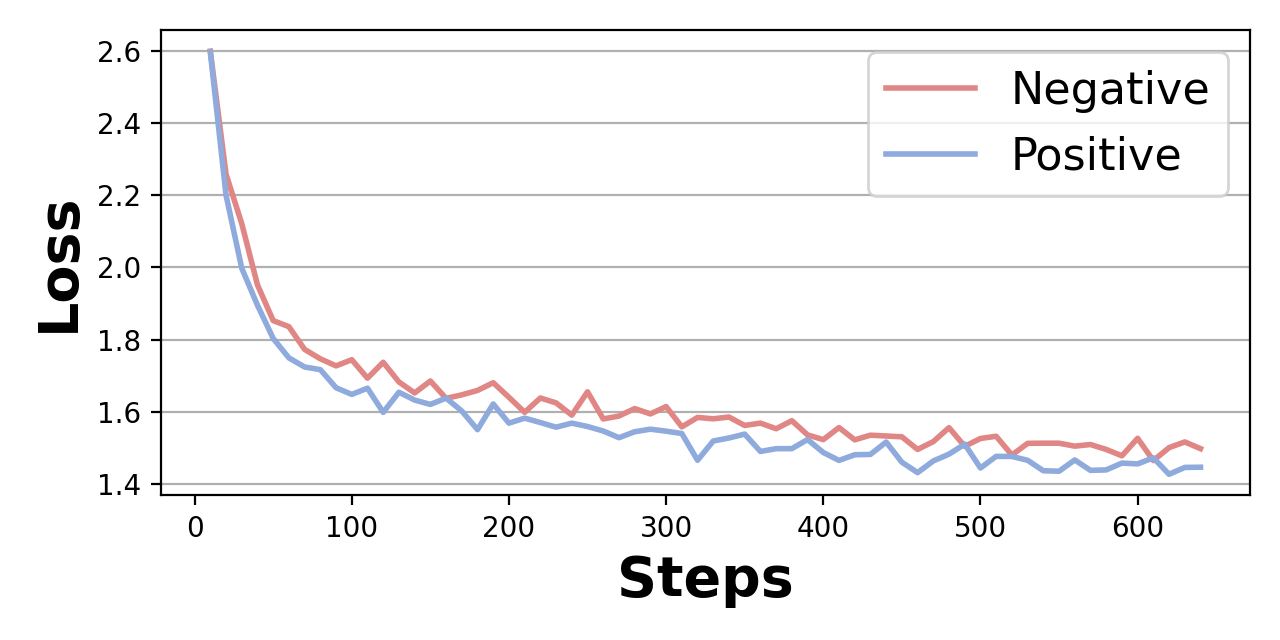} 
\end{subfigure}
\begin{subfigure}{.475\textwidth}
  \centering
   \includegraphics[width=\linewidth]{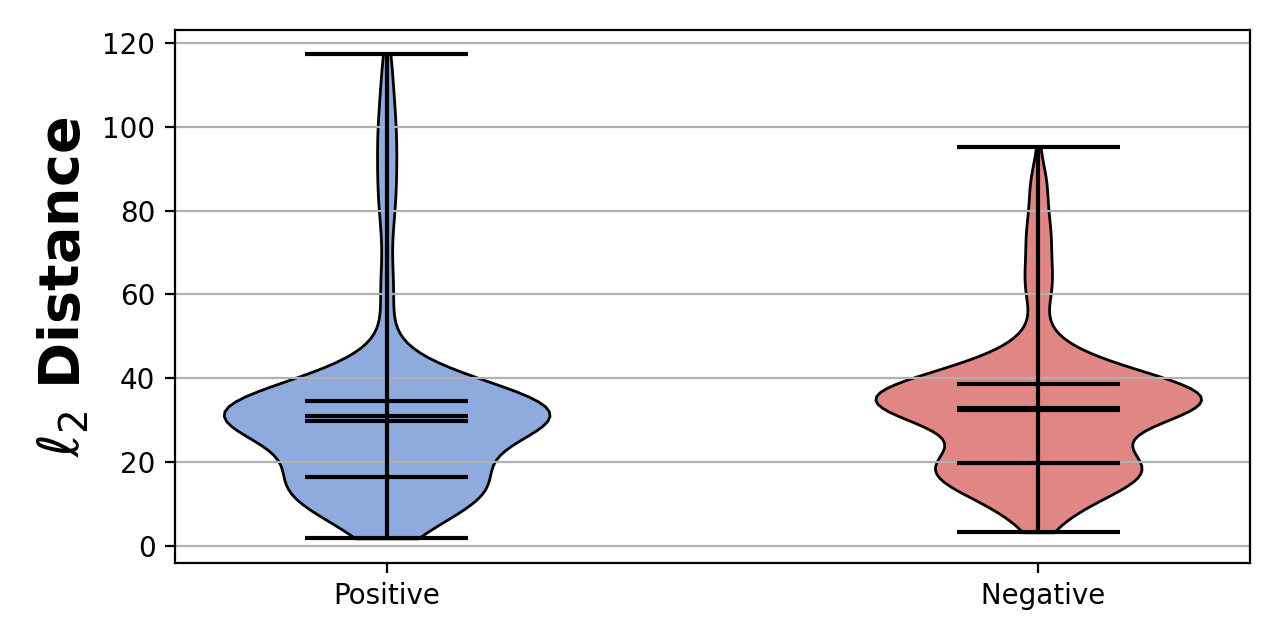}  
\end{subfigure}
\caption{Pre-training Loss of Positive and Negative Models (Left) and Violin Plots of $\ell_2$ Distance for Positive and Negative Representation (Right).}
\label{fig:pretrain}
\end{figure}

To demonstrate the effect of the differences between positive and negative representations, we randomly sampled 500 triples from the contrastive dataset and recorded the $\ell_2$ distance between the original representation and the positive representation, as well as the original representation and the negative one.
%
We present the statistics of the recorded 500 pairs of $\ell_2$ distances in Table~\ref{tab:dist_stats}, with $\mathcal{L}^{+}$ and $\mathcal{L}^{-}$ defined in Eq.~\eqref{eqn:l2_dist_p} and Eq.~\eqref{eqn:l2_dist_n}, respectively. The row labeled `Mean' denotes the average $\ell_2$ distance, while the row labeled `Std.' indicates the standard deviation of the $\ell_2$ distance values. Furthermore, we visualize the data distribution in Figure~\ref{fig:pretrain} (Lower).
%
From the visualization, we observe that the $\ell_2$ norms of positive and negative representations show significant differences in terms of their distributions. Furthermore, we compute the KL divergence $KL(P,N) = \sum_{x \in \mathcal{X}}P(x) \log \left(P(x)/N(x)\right)$ between the two $\ell_2$-norm vectors to verify this observation.


\subsection{TruthfulQA Evaluation (RQ.2)}
We present the evaluation results of the TruthfulQA Benchmark~\cite{lin-etal-2022-truthfulqa} for measuring the hallucination of trained models in Table~\ref{tab:hallucination}. From Table~\ref{tab:hallucination}, it is evident that our method, \modelname, exhibits a significant improvement in the MC1 score based on the TruthfulQA evaluation compared to both the foundation models and the existing \textbf{LoRRA}~\cite{zou2023representation} method. Specifically, for the \textbf{LLaMA2} foundation model, we improve the MC1 score by up to 19.43 points compared to the foundation model and by 3.82 points compared to \textbf{LoRRA}. A consistent improvement can also be observed for the \textbf{Alpaca} and \textbf{LLaMA3} models. Notably, for the \textbf{Alpaca} model, the \textbf{LoRRA} method decreases the MC1 score, while our method increases it by up to 3.38 points. In conclusion, \modelname~demonstrates efficient and consistent improvement in the hallucination reduction task across the three foundation models.

\subsection{Knowledge Evaluation (RQ.3)}
To answer RQ.3, we present the knowledge evaluation of the model fine-tuned with our method, \modelname, in comparison to the foundation models and the \textbf{LoRRA} models based on the benchmarks of the \textbf{MMLU}~\cite{hendrycks2020measuring} and \textbf{C-Eval}~\cite{huang2024c} datasets. Both \textbf{MMLU} and \textbf{C-Eval} are designed to evaluate the model's performance across a wide range of subjects. \textbf{MMLU} focuses on multiple-choice questions in various academic disciplines, while \textbf{C-Eval} targets a comprehensive set of tasks that include both multiple-choice and open-ended questions to assess the models' capabilities in different linguistic and knowledge domains. The results for \textbf{MMLU} and \textbf{C-Eval} are presented in Figure~\ref{fig:radar}. We observe that \modelname~exerts no significant negative influence on the knowledge evaluation, demonstrating the model capability-preserving property.

\begin{table}[t!]
\centering
\caption{Statistics of $\ell_2$ Distance and its KL Divergence.}
\begin{tabular}{c c c } 
\hline 
 & $\mathcal{L}^{+}$ & $\mathcal{L}^{-}$ \\
 \hline 
Mean & 30.9388 &  32.4959\\
Std. & 21.1840 &  16.3295 \\
& $KL(P, N)$ & $KL(N, P)$ \\
KL Divergence & 0.0163 & 0.0162 \\
 \hline 
\end{tabular}
\label{tab:dist_stats} 
\end{table}

\vspace{-1.0pt}

\begin{table}[t!]
\centering
\caption{MC1 Score with TruthfulQA Evaluation~\cite{lin-etal-2022-truthfulqa} of \modelname~Compared to Foundation Models and LoRRA~\cite{zou2023representation}.}
\begin{tabular}{c  c c  c } 
\hline 
Methods & \textbf{LlaMA2}~\cite{touvron2023llama} & \textbf{Alpaca}~\cite{alpaca} & \textbf{LLaMA3}~\cite{dubey2024llama}\\
 \hline 
\textbf{Foundation} & 0.3185 & 0.2807 & 0.2166 \\
\textbf{LoRRA}~\cite{zou2023representation} & \underline{0.4736} &  \underline{0.2337} & \underline{0.2680} \\
\modelname & \textbf{0.5128} & \textbf{0.3145} & \textbf{0.2705} \\
 \hline 
\end{tabular}
\label{tab:hallucination} 
\end{table}


\begin{figure*}[htbp]
   \centering
   \includegraphics[width=1.0\textwidth]{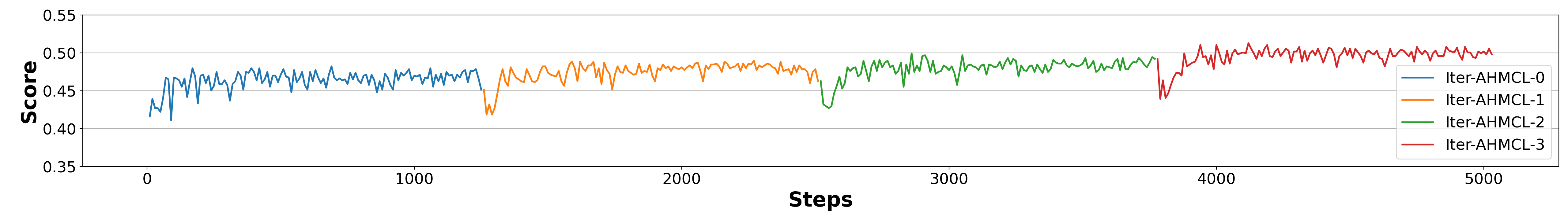}
   \caption{Iterative Process of Model Guidance on Foundation Model \textbf{LLaMA2}.}
   \label{fig:iter_process}
\end{figure*}

\subsection{Model Evaluation by QWen API(RQ.3)}
\label{sec:qwen}

In this section, we discuss the use of the QWen LLM API for evaluation~\cite{bai2023qwen}. The process involves utilizing foundational or fine-tuned models to generate responses for CNN-DailyMail News Text Summarization~\cite{chen2017examination}. Answer sheets are created from the model outputs and standardized answers, which are then input into the Qwen API. The API evaluates the answer sheets and provides results, which are subsequently analyzed statistically. The evaluation encompasses four aspects: Relevance, Fluency, Coherence, and Consistency.

We present the evaluation results in Table~\ref{tab:qwen_results}, based on the Qwen API. Our ongoing analysis of the Qwen scores, as shown in Table~\ref{tab:qwen_results}, begins by noting that the \textbf{Gold-Ref} achieves the highest scores across all four perspectives, reflecting the tailored evaluation methodology of Qwen. Furthermore, our method, \modelname, consistently delivers stable results in all evaluation dimensions, slightly outperforming the \textbf{Foundation} models but not reaching the level of the \textbf{Gold-Ref}.

\begin{table}[htbp]
\centering
\caption{Qwen API~\cite{bai2023qwen} Evaluation Results.}
\label{tab:qwen_results}
\resizebox{0.48\textwidth}{!}{
\begin{tabular}{c cccc} 
\toprule
Methods & Relevance $\uparrow$ & Fluency $\uparrow$ & Coherence $\uparrow$ & Consistency $\uparrow$ \\
\midrule
& \multicolumn{4}{c}{\textbf{LLaMA2}~\cite{touvron2023llama}} \\
\textbf{Gold-Ref} & \textbf{3.84} $\pm$ 0.44 & \textbf{4.97} $\pm$ 0.17 & \textbf{3.96} $\pm$ 0.34 & \textbf{4.69} $\pm$ 0.48 \\
\textbf{Foundation} & 2.51 $\pm$ 0.73 & 3.38 $\pm$ 0.90 & 2.34 $\pm$ 0.76 & 2.94 $\pm$ 0.98\\
\textbf{LoRRA} & 2.55 $\pm$ 0.74 & 3.50 $\pm$ 0.95 & 2.41 $\pm$ 0.83 & 2.99 $\pm$ 1.02 \\
\textbf{\modelname} & \underline{2.60} $\pm$ 0.73 & \underline{3.50} $\pm$ 0.92 & \underline{2.44} $\pm$ 0.80 & \underline{3.10} $\pm$ 0.98\\
\midrule 
& \multicolumn{4}{c}{\textbf{Alpaca}~\cite{alpaca}} \\
\textbf{Gold-Ref} & \textbf{3.91} $\pm$ 0.49 & \textbf{4.97} $\pm$ 0.17 & \textbf{4.04} $\pm$ 0.34 & \textbf{4.75} $\pm$ 0.46 \\
\textbf{Foundation} & 2.47 $\pm$ 0.84 & 3.10 $\pm$ 1.04 & 2.20 $\pm$ 0.77 & 2.96 $\pm$ 1.06 \\
\textbf{LoRRA} & 2.41 $\pm$ 0.78 & 3.08 $\pm$ 1.05 & 2.17 $\pm$ 0.75 & 2.92 $\pm$ 1.06 \\
\textbf{\modelname} & \underline{2.58} $\pm$ 0.80 & \underline{3.29} $\pm$ 1.07 & \underline{2.34} $\pm$ 0.84 & \underline{3.12} $\pm$ 1.06\\
\midrule 
& \multicolumn{4}{c}{\textbf{LLaMA3}~\cite{dubey2024llama}} \\
\textbf{Gold-Ref} & \textbf{3.78} $\pm$ 0.52 & \textbf{4.98} $\pm$ 0.14 & \textbf{3.99} $\pm$ 0.30 & \textbf{4.68} $\pm$ 0.49 \\
\textbf{Foundation} & \underline{2.57} $\pm$ 0.74 & \underline{3.60} $\pm$ 0.69 & \underline{2.33} $\pm$ 0.71 & \underline{3.21} $\pm$ 0.90 \\
\textbf{LoRRA} & 2.55 $\pm$ 0.77 & 3.59 $\pm$ 0.74 & 2.31 $\pm$ 0.73 & 3.18 $\pm$ 0.94 \\
\textbf{\modelname} & 2.53 $\pm$ 0.75 & 3.58 $\pm$ 0.74 & 2.30 $\pm$ 0.71 & 3.17 $\pm$ 0.91\\
\bottomrule
\end{tabular}}
\end{table}


\begin{table*}[htbp]
\centering
\caption{Benefits of Iterative Model Guidance with \textbf{Alpaca} as Foundation Model.}
\label{tab:iterative}
\resizebox{0.9\textwidth}{!}{
\begin{tabular}{cccc ccc} 
\toprule
$i$ &  $\mathcal{M}^{+}_{i}$ & $\mathcal{M}^{+}_{i-1}$ & $\mathcal{M}^{-}$ & \textbf{LLaMA2}~\cite{touvron2023llama} & \textbf{Alpaca}~\cite{alpaca} & \textbf{LLaMA3}~\cite{dubey2024llama}\\
\midrule
0 & \textbf{Foundation} & $\varnothing$ & $\varnothing$ & 0.3145 & 0.2007 &  0.2166 \\
1 & \textbf{\modelname-0} & \textbf{GMP-Positive} & \textbf{GMP-Negative} & 0.4736 & 0.2447  & 0.2582\\
2 & \textbf{\modelname-1} & \textbf{\modelname-0} & \textbf{GMP-Negative} & 0.4810& 0.2509 &  \underline{0.2680} \\
3 & \textbf{\modelname-2} & \textbf{\modelname-1} & \textbf{GMP-Negative} & \underline{0.4908} & \underline{0.2582}& \textbf{0.2705} \\
4 & \textbf{\modelname-3} & \textbf{\modelname-2} & \textbf{GMP-Negative} & \textbf{0.5128}& \textbf{0.3145} & \textbf{0.2705} \\
\midrule 
1 & \textbf{\modelname-Transfer} & \textbf{\modelname-4-LLaMA2} & \textbf{GMP-Negative} & - & 0.2472 & -  \\
\bottomrule
\end{tabular}
}
\end{table*}

\subsection{Iterative Process of Model Guidance (RQ.4)}

\begin{figure}[t!]
  \centering
   \includegraphics[width=0.6\linewidth]{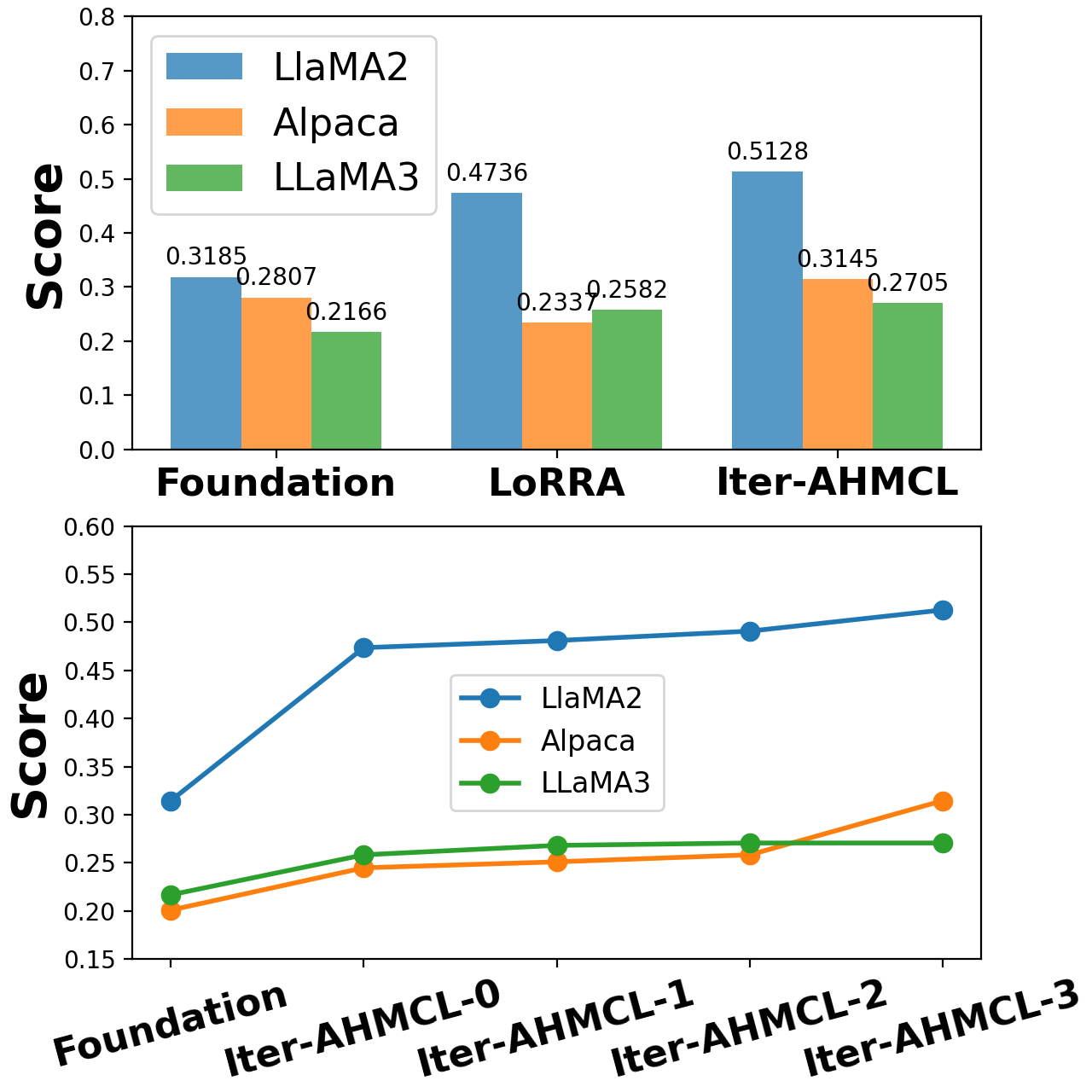}  
\caption{Bar Compare of TruthfulQA~\cite{lin-etal-2022-truthfulqa} (Upper) and Checkpoints Evaluation during \modelname ~(Lower).}
\label{fig:check}
\end{figure}

This section demonstrates the benefits of iterative model guidance in addressing \textbf{RQ.4}. We take \modelname~applied to the foundation model \textbf{Alpaca} as an example. The training procedure is described in Algorithm~\ref{alg:iter-ahmcl}. All training is conducted with a maximum of 1,250 iteration steps.
The detailed iterative process of \modelname~and its improvements on the TruthfulQA Evaluation are shown in Table~\ref{tab:iterative}. By updating the positive guidance model, we iteratively enhance the MC1 score of the fine-tuned model. The long-term iterative process is illustrated in Figure~\ref{fig:iter_process}, where the x-axis represents the training steps and the y-axis represents the MC1 score evaluated using TruthfulQA. From Figure~\ref{fig:iter_process}, we observe that the score improves progressively, with oscillations occurring at the interchange points of the guidance model. The iterative updates of the positive guidance model may temporarily degrade the model's performance around the turning points, but they ultimately contribute to long-term improvements in the MC1 score. Additionally, we present the evaluation of checkpoints for three foundation models trained with \modelname~in Figure~\ref{fig:check} (Lower), which demonstrates consistent improvements across different foundation models.

\subsection{Transferability (RQ.5)}
In this section, we address \textbf{RQ.5}: whether a pre-trained guidance model based on one foundation model can be transferred for the \textbf{CL-MG}/\textbf{CL-IMG} learning to another foundation model. To explore this, we conducted experiments using a positive guidance model trained on the foundation model \textbf{LLaMA2} and \modelname~trained on \textbf{Alpaca}. The TruthfulQA Evaluation MC1 scores are presented in the last row of Table~\ref{tab:iterative}. In this experiment, the positive guidance model used is \textbf{\modelname-LLaMA2} at iteration 6, with the negative guidance model remaining the same as in other experiments. The transfer experiments yield performance comparable to the first iteration of \modelname~using a positive guidance model tuned from a homogeneous guidance model, demonstrating the transferability.

\section{Conclusion}

In our paper, we introduce a novel method called \modelname, designed to reduce hallucination in LLM models while preserving their overall performance. This approach involves fine-tuning the representations at specific layers to enhance the model's capabilities through constructive learning. Unlike existing strategies that rely on sample-level contrasts, we propose formulating guidance at the feature representation level using specifically trained positive and negative model guidance. This allows us to establish contrasts at both the sample and model levels. Furthermore, this contrastive learning approach can be conducted iteratively by continuously updating the positive guidance model.
Our comprehensive experiments, which include evaluations of hallucination and language ability, demonstrate the efficiency and effectiveness of our proposed methods.
In the future, we plan to investigate the transferability of \modelname~in the cross-domain scenarios.

\bibliographystyle{plain}
\bibliography{llm_are}


\end{document}